\documentclass[a4paper,fleqn]{cas-dc}
\usepackage{dblfnote}
\usepackage[sort,numbers]{natbib}
\usepackage{hyperref}
\usepackage{algorithm}
\usepackage{algorithmic}
\usepackage{multicol}
\usepackage[caption=false,font=normalsize,labelfont=sf,textfont=sf]{subfig}
\usepackage{amsmath,amsfonts}
\usepackage{algorithmic}
\usepackage{algorithm}
\usepackage{array}
\usepackage{textcomp}
\usepackage{stfloats}
\usepackage{url}
\usepackage{verbatim}
\usepackage{graphicx}
\usepackage{algorithm, bm}
\usepackage{algorithmic}
\usepackage{booktabs}
\usepackage{multirow}
\usepackage{makecell}
\usepackage{times}
\usepackage{pifont}
\usepackage{color}

\usepackage{subfig,graphicx}
\usepackage[T1]{fontenc}
\usepackage{mathdesign}
\usepackage{lipsum}
\usepackage{titlesec}
\newcommand{\charternumbers}{\fontfamily{bch}\selectfont}
\DeclareTextFontCommand{\textcharter}{\charternumbers}

\titleformat{\section}{\normalfont\fontsize{9pt}{12pt}\bfseries}{\textcharter{\thesection.}}{0.5em}{}
\titleformat{\subsection}{\normalfont\fontsize{8pt}{12pt}\itshape}{\textcharter{\thesubsection.}}{0.3em}{}
\titleformat{\subsubsection}{\normalfont\fontsize{8.3pt}{12pt}\itshape}{\textcharter{\thesubsubsection.}}{0.4em}{}

\titleformat{name=\section,numberless}{\normalfont\fontsize{9pt}{12pt}\bfseries}{}{0em}{}
\titleformat{name=\subsection,numberless}{\normalfont\fontsize{9pt}{12pt}\bfseries}{}{0em}{}
\titleformat{name=\subsubsection,numberless}{\normalfont\fontsize{9pt}{12pt}\bfseries}{}{0em}{}

\usepackage{hyperref}

\usepackage{balance}
\usepackage{color}
\hypersetup{
  colorlinks=true,
  linkcolor=cyan,
  urlcolor=cyan,
  citecolor=cyan,
  linktoc=all
}
\usepackage{fancyhdr}
\def\tsc#1{\csdef{#1}{\textsc{\lowercase{#1}}\xspace}}
\tsc{WGM}
\tsc{QE}
\tsc{EP}
\tsc{PMS}
\tsc{BEC}
\tsc{DE}

\begin{document}
\thispagestyle{empty}

\let\WriteBookmarks\relax
\def\floatpagepagefraction{1}
\def\textpagefraction{.001}
\let\printorcid\relax

\shorttitle{Leveraging social media news}


\title[mode = title]{Exploring the Adversarial Robustness of Face Forgery Detection with Decision-based Black-box Attacks}  



\author[1,2,3]{\textcolor{black}{Zhaoyu Chen}}[type=editor,
    auid=000,bioid=2,
    role=,]
\ead{zhaoyuchen20@fudan.edu.cn} 

\author[4]{\textcolor{black}{Bo Li}}[style=chinese]
\ead{njumagiclibo@gmail.com}

\author[1,2]{\textcolor{black}{Kaixun Jiang}}[type=editor,
    auid=000,bioid=2,
    role=,]
\ead{kxjiang22@m.fudan.edu.cn} 

\author[3]{\textcolor{black}{Shuang Wu}}[style=chinese]
\ead{calvinwu@tencent.com}

\author[3]{\textcolor{black}{Shouhong Ding}}[style=chinese]
\ead{ericshding@tencent.com}

\author[1,2,5]{\textcolor{black}{Wenqiang Zhang}}
\cormark[1]
\ead{wqzhang@fudan.edu.cn} 

\address[1]{Shanghai Engineering Research Center of AI \& Robotics, Academy for Engineering \& Technology, Fudan University, Shanghai 200433, China}
\address[2]{Engineering Research Center of Robotics, Ministry of Education, Academy for Engineering \& Technology, Fudan University, Shanghai 200433, China}
\address[3]{Youtu Lab, Tencent, Shanghai 200000, China}
\address[4]{vivo Mobile Communication Co., Ltd, Shanghai 200120, China}
\address[5]{Shanghai Key Lab of Intelligent Information Processing, School of Computer Science, Fudan University, Shanghai 200433, China}

\cortext[1]{Corresponding author} 

\begin{abstract}
Face forgery generation technologies generate vivid faces, which have raised public concerns about security and privacy. Many intelligent systems, such as electronic payment and identity verification, rely on face forgery detection. Although face forgery detection has successfully distinguished fake faces, recent studies have demonstrated that face forgery detectors are very vulnerable to adversarial examples. Meanwhile, existing attacks rely on network architectures or training datasets instead of the predicted labels, which leads to a gap in attacking deployed applications. To narrow this gap, we first explore the decision-based attacks on face forgery detection. 
We identify challenges in directly applying existing decision-based attacks, such as perturbation initialization failure and reduced image quality. To overcome these issues, we propose cross-task perturbation to handle initialization failures by utilizing the high correlation of face features on different tasks. Additionally, inspired by the use of frequency cues in face forgery detection, we introduce the frequency decision-based attack. This attack involves adding perturbations in the frequency domain while constraining visual quality in the spatial domain.
Finally, extensive experiments demonstrate that our method achieves state-of-the-art attack performance on FaceForensics++, CelebDF, and industrial APIs, with high query efficiency and guaranteed image quality. Further, the fake faces by our method can pass face forgery detection and face recognition, which exposes the security problems of face forgery detectors.
\end{abstract}
\begin{keywords}
Adversarial examples \sep Face forgery detection \sep Black-box attacks \sep Face recognition \sep Decision-based attacks
\end{keywords}

\maketitle


\thispagestyle{empty}

\section{Introduction}
\label{sec:intro}

With the success of deep neural networks~\cite{resnet,xception,EfficientNet,vit,chen2022towards} and the development of generative models~\cite{DBLP:conf/nips/GoodfellowPMXWOCB14,ddim,beatgan,ldm}, face forgery generation has made excellent progress. We can synthesize highly realistic fake faces by these techniques, such as Face2Face~\cite{Face2Face}, FaceSwap~\cite{Faceswap}, DeepFakes~\cite{Deepfake} and NeuralTextures~\cite{neuraltextures}. Further, methods such as StyleGAN~\cite{stylegan} edit one’s attributes (hair, glasses, age, etc.) to generate various faces, and these faces are widely used in scenes such as the film industry and virtual uploaders. In contrast, these fake faces also lead to the spread of false news or damage to reputation, raising public concerns about security and privacy~\cite{sun2023contrastive,fas1,fas2,fas3,fas4,fas5}. Therefore, researchers have recently focused on how to design effective face forgery detection methods to determine whether a face has been modified.

\begin{figure}
    \centering
    \includegraphics[width=0.47\textwidth]{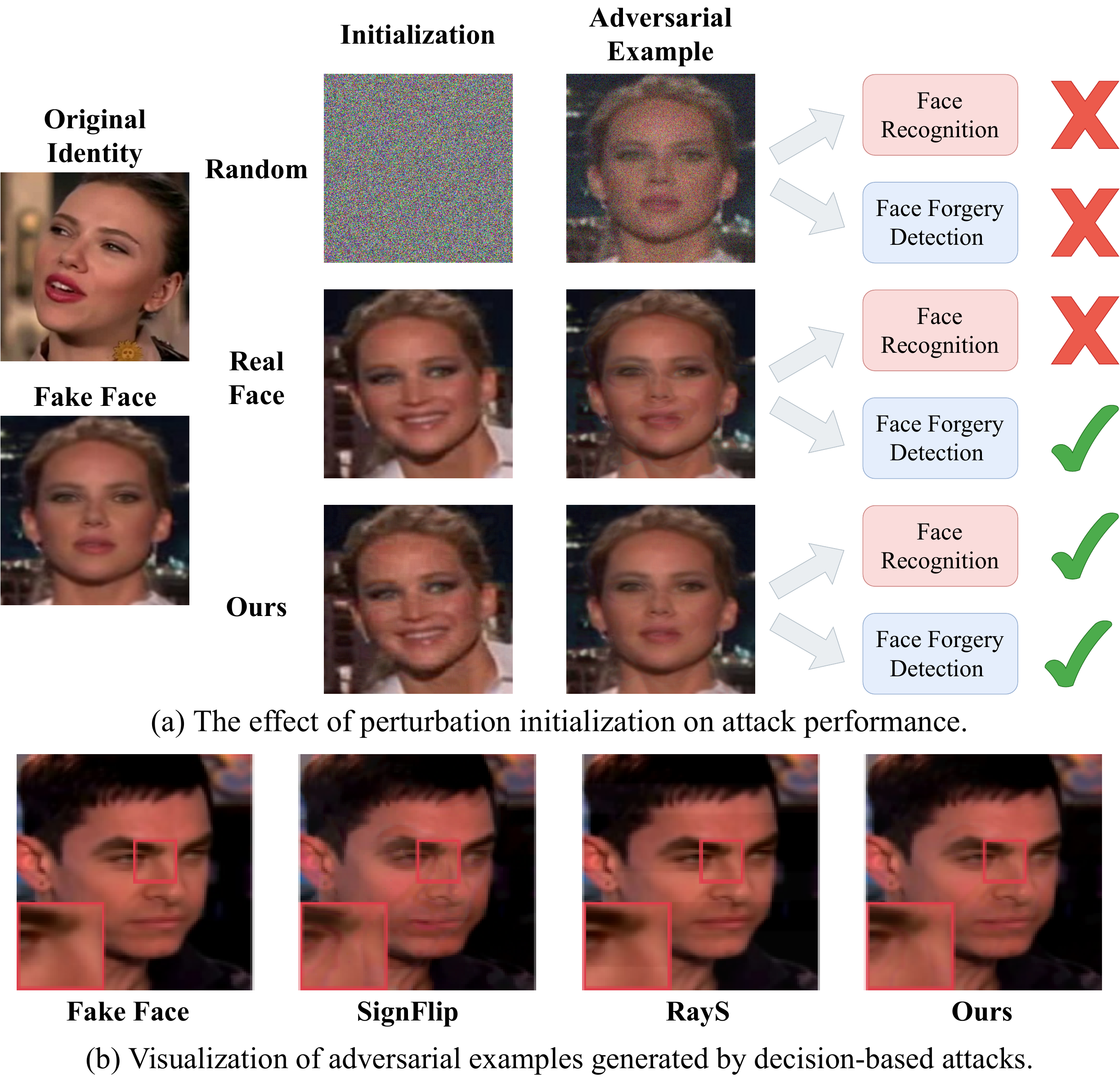}
    \caption{When directly using existing decision-based attacks on face forgery detection, it is prone to (a) \textit{perturbation initialization failure} and (b) \textit{low image quality}, which affects attack performance and limits the application of the face (i.e. simultaneously attacking face recognition and face forgery detection). {\color{OliveGreen}{\ding{52}}} represents recognition as the original identity or detection as real faces.} 
    \label{fig:bypass}
\end{figure} 

Face forgery detection is usually defined as a binary classification problem, distinguishing between real and fake faces. There is a series of face forgery detectors~\cite{resnet,xception,EfficientNet,vit,f3net,M2TR} proposed that utilize deep neural networks to automatically learn feature differences between real and fake faces and achieve remarkable success on multiple datasets~\cite{ffpp,CelebDF}. Despite their success, recent work~\cite{EvadingDF} shows that face forgery detectors are very vulnerable to adversarial examples~\cite{chen2022shape,fgsm,devopatch}, which exposes the current shortcut of security in face forgery detection. After adding a carefully-designed adversarial perturbation, a face originally classified as the fake is recognized as real by face forgery detectors. Existing methods~\cite{EvadingDF,KeyRegionAttack,AdvDfPractical,FaceManifold,hybridfre,huang2022cmua} have explored the adversarial robustness of face forgery detection, but these methods require access to the detector's network architecture or training datasets. 
When targeting industrial APIs in the context of face forgery detection, we do not have access to the underlying models or datasets. Instead, we can only observe the binary decision (real or fake) provided by the service. Consequently, the absence of decision-based attacks results in a significant gap in the robustness evaluation of face forgery detection systems.

When exploring decision-based attacks on face forgery detection, it is easiest to apply existing decision-based attacks directly~\cite{boundary,OptAttack,signopt, hsja,SignFlip,TA}. However, directly applying existing decision-based attacks to face forgery detection may encounter some issues: 
(a) \textit{perturbation initialization failure}. Existing decision-based attacks usually adopt random noise as attack initialization. However, since adding large noise itself is a kind of manipulation, faces with large random noise are usually recognized as fake faces by face forgery detection (as shown in Figure~\ref{fig:bypass}). 
(b) \textit{low image quality}. For example, RayS~\cite{RayS} introduces perceptible regular rectangular perturbation. 
Using real faces for perturbation initialization~\cite{SignFlip} can introduce artifacts and potentially impact face-related applications, such as face recognition. Even if such faces succeed in bypassing forgery detection models, they fail to achieve the ultimate goal of deceiving face recognition systems, as illustrated in Figure~\ref{fig:bypass}.
Therefore, we need an efficient decision-based attack designed for face forgery detection to identify potential vulnerabilities on face forgery detection and provide insights on possible improvements.

To address these issues, we propose an efficient decision-based adversarial attack to explore the adversarial robustness of face forgery detection. First, we propose a cross-task perturbation initialization method to handle initialization failures of decision-based attacks on face forgery detection. 
With the high correlation of face features on face forgery detection and face recognition, we generate cross-task adversarial perturbations in the face recognition model as initial perturbations to attack face forgery detection. Second, since the features of face forgery detection are highly discriminative in the frequency domain~\cite{fre_analysis,f3net,M2TR}, we propose a frequency decision-based attack. After obtaining cross-task adversarial perturbations, we add perturbations to faces in the frequency domain and then constrain the visual quality in the spatial domain. Finally, our method achieves state-of-the-art attack performance on FaceForensics++~\cite{ffpp} and CelebDF~\cite{CelebDF}, with high query efficiency and guaranteed image quality. We also successfully attack the real-world industrial API and showcase the effectiveness of our method in practice. 
Furthermore, fake faces by our method can simultaneously pass face forgery detection and face recognition, which exposes the current adversarial vulnerability of face-related systems.

Our main contributions can be summarized as follows:

\begin{itemize}
    \item We first explore decision-based attacks on face forgery detection and verify their adversarial vulnerability in academic and industrial scenarios. Fake faces by our method can pass face forgery detection and face recognition, which further exposes the security problems of face forgery detectors.
    \item We propose cross-task perturbation initialization to handle initialization failures to attacks by leveraging the high correlation of face features across tasks without requiring any knowledge of detectors.
    \item We propose a frequency decision-based attack based on frequency clues, which for the first time verifies the effectiveness of frequency domain-based query attacks on face forgery detection.
    \item Experiments show that our method achieves state-of-the-art attack performance on FaceForensics++ and CelebDF, and first show that frequency-based detectors are just as vulnerable as spatial-based ones.
\end{itemize}

The remainder of the paper is organized as follows. Section~\ref{sec:relatedwork} briefly introduces face forgery detection and decision-based black-box attacks and then discusses adversarial attacks on face forgery detection. Section~\ref{sec:method} first gives the formal definition of decision-based attacks on face forgery detection. Then, we introduce Cross-task Perturbation Initialization and propose the Frequency Decision-based Attack. Section~\ref{sec:experiments} presents experiments to illustrate the effectiveness of the proposed method, including attacks on state-of-the-art detectors, ablation studies, attacks on industrial APIs, etc. We further discuss the possibility of using face-related tasks to assist in attacking face forgery detection in Section~\ref{sec:discussion} and conclude this paper in Section~\ref{sec:conclusions}.

\section{Related Work}
\label{sec:relatedwork}

\subsection{Face Forgery Detection}
Since large-scale public datasets are freely available, combined with rapid progress in generative models, a very realistic fake face has been generated with its corresponding implications for society. Therefore, there have been many efforts to distinguish real and fake faces as much as possible in face forgery detection. Early work~\cite{ffpp} directly uses convolutional neural networks (CNNs) to extract discriminative features for face forgery detection. However, these spatial features extracted by CNNs focus more on class-level distinctions rather than subtle differences between real and fake faces. Furthermore, recent work~\cite{fre_analysis,f3net,M2TR} mines forgery patterns and proposes many frequency-based detection methods, noting that real and fake faces are diverse in the frequency domain. As a two-branch network, F$^3$Net~\cite{f3net} finds forgery patterns using frequency clues and extracts discrepancies between real and fake images by looking at frequency statistics. Considering the potential of the recent Vision Transformer~\cite{vit}, M2TR~\cite{M2TR} utilizes a multi-scale transformer for capturing local inconsistencies at different scales and additionally introduces frequency modality to improve performance with multiple image compression algorithms. 
In addition, recent works have utilized self-augmentation~\cite{sbi}, attention~\cite{fed2023,nguyen2024laa}, multimodality~\cite{modalffd}, and mixture-of-experts~\cite{moeffd} to further obtain more generalized detectors.
To better evaluate the adversarial robustness of face forgery detection, we choose more classic and more generalizable spatial-based~\cite{resnet,xception,EfficientNet,vit} and frequency-based detectors~\cite{f3net,M2TR} for experiments.

\subsection{Decision-based Black-box Attacks}
In contrast to white-box attacks that require access to model details such as architecture and gradients, recent work has focused on black-box attacks~\cite{boundary,chen2023aca,Jiang_2023_ICCV,10.1145/3581783.3611828,WANG2024124757,10.1145/3581783.3611827}, including transfer-based, score-based, and decision-based attacks. Transfer-based attacks need to know the distribution of the data to construct the surrogate model and its performance is limited by the surrogate model. The difference between score-based and decision-based attacks is that the output returned by the threat model is a confidence or a label. It is obvious that the decision-based setting is the most difficult, but it is also the most practical since in reality, the adversary only knows minimal information about the model.

Brendel et al.~\cite{boundary} propose the first decision-based attack to randomly walk on the decision boundary starting from an adversarial point while keeping adversarial, called the Boundary Attack. Some works~\cite{OptAttack,signopt,TA,hsja} improve Boundary Attack from the perspective of gradient optimization. Next, SignFlip~\cite{SignFlip} introduces noise projection and random sign flip to improve decision-based attack by large margins. RayS~\cite{RayS} reduces the number of queries by reformulating the continuous problem of finding the closest decision boundary into a discrete problem, which does not require zeroth-order gradient estimation. In this paper, we explore decision-based attacks on face forgery detection under $l_\infty$ norm, i.e. adding perturbation within the perturbation range $\epsilon$, which is more practical in real-world applications.

\subsection{Adversarial Attacks on Face Forgery Detection}
Face forgery detection is a security-sensitive task and its adversarial vulnerability has received much attention~\cite{EvadingDF,KeyRegionAttack,AdvDfPractical,FaceManifold,hybridfre}. Carlini et al.~\cite{EvadingDF} and Neekhara et al.~\cite{AdvDfPractical} shows the adversarial robustness of face forgery detectors under various white-box attacks and transfer-based attacks. Hussain et al.~\cite{advdeepfake} evaluate the detectors by the score-based attack with natural evolutionary strategies~\cite{nes}. Li et al.~\cite{FaceManifold} manipulate the noise and latent vectors of StyleGAN~\cite{stylegan} to fool face forgery detectors. KRA~\cite{KeyRegionAttack} attacks the key regions of forged faces obtained from the semantic region selection and boosts the adversarial transferability. Then, Jia et al.~\cite{hybridfre} add adversarial perturbations in the frequency domain and generate more imperceptible adversarial examples with a high transferability. However, existing work is unable to evaluate adversarial robustness in a decision-based setting.
In this paper, considering academic and industrial applications, we focus on attacking frame-level detectors because recent attacks on face forgery detection have mainly focused on attacking frame-level detectors~\cite{EvadingDF,KeyRegionAttack,AdvDfPractical,FaceManifold,hybridfre}, and industrial APIs tend to use images as input because of the low computation and communication costs. 
In conclusion, we evaluate the adversarial robustness of face forgery detection under decision-based attacks for the first time and hope to provide insight into possible future improvements.

\section{Methodology}
\label{sec:method}
In this section, we first introduce preliminaries about decision-based attacks and the threat model. Then, we propose the cross-task perturbation initialization. Finally, we introduce the frequency perturbation into the decision-based attacks against face forgery detection.

\subsection{Preliminaries}
We define a neural network-based face forgery detector $f: \mathbb{R}^d \rightarrow \mathbb{R}^k$ as the threat model. Because the face forgery detection is a binary classification task, its label $y$ is real or fake, i.e. $y \in \mathbb{R}^2 = \{0,1\}$. Given a face $x \in [0,1]^d$, we define $f(x)_i$ as the probability of class $i$ and $c(x) = \mathrm{argmax}_{i \in \{0,1\}}\ f(x)_i$ refers to the predicted label. In this paper, we only consider faces that are classified as fake and treat the situation where fake faces are classified as real faces as successful attacks. Therefore, the adversary aims to find an adversarial perturbation $\delta \in \mathbb{R}^d$ such that $c(x+\delta)=0$ (classified as real). Here, we define $||\delta||_{\infty} \leq \epsilon$  and $\epsilon$ is the allowed maximum perturbation. 

In the decision-based setting, the adversary does not have access to the model's network architecture, weights, gradients, or the model's predicted probabilities $f(x)$, only the model's predicted labels $c(x)$. Therefore, the optimization objective of the decision-based attack is followed as:
\begin{equation}
    \label{eq:objective}
    \min_{\delta} ||\delta||_{\infty}\quad\mathrm{s.t.}\quad c(x+\delta)=0.
\end{equation}
The adversary keeps querying the model and optimizing adversarial perturbations $\delta$ until $||\delta||_{\infty} \leq \epsilon$.

\begin{figure}[t]
    \begin{center}
    \includegraphics[width=0.5\textwidth]{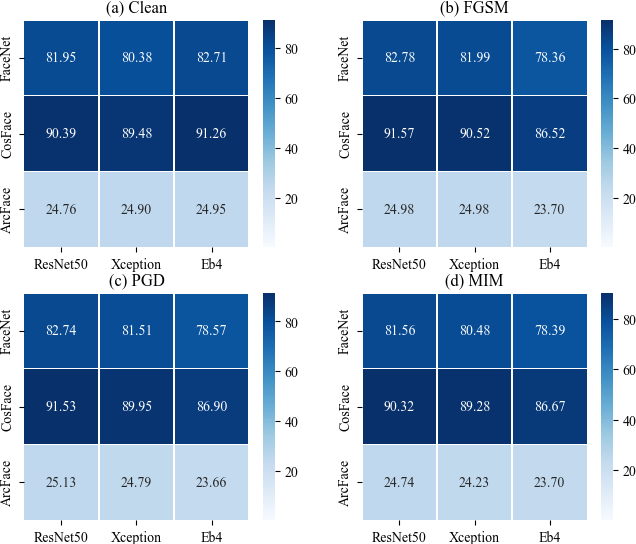}
    \end{center}
    \caption{Empirical analysis of cosine similarity (\%) on intermediate features between forgery detectors and face recognition models under clean examples and adversarial examples (FGSM~\cite{fgsm}, PGD~\cite{pgd}, and MIM~\cite{mim}). Here, the face recognition models are FaceNet~\cite{FaceNet}, CosFace~\cite{CosFace}, and ArcFace~\cite{ArcFace}. The face forgery detection models are ResNet50~\cite{resnet}, Xception~\cite{xception}, EfficientNet-b4 (Eb4)~\cite{EfficientNet}. We find that face recognition models and face forgery detectors have a high correlation between the intermediate features of the same face. It means that the perturbation of attacking face recognition can also perturb face forgery detection to a certain extent.}
    \label{fig:feature}
\end{figure}

\begin{figure*}[t]
    \begin{center}
    \includegraphics[width=1\textwidth]{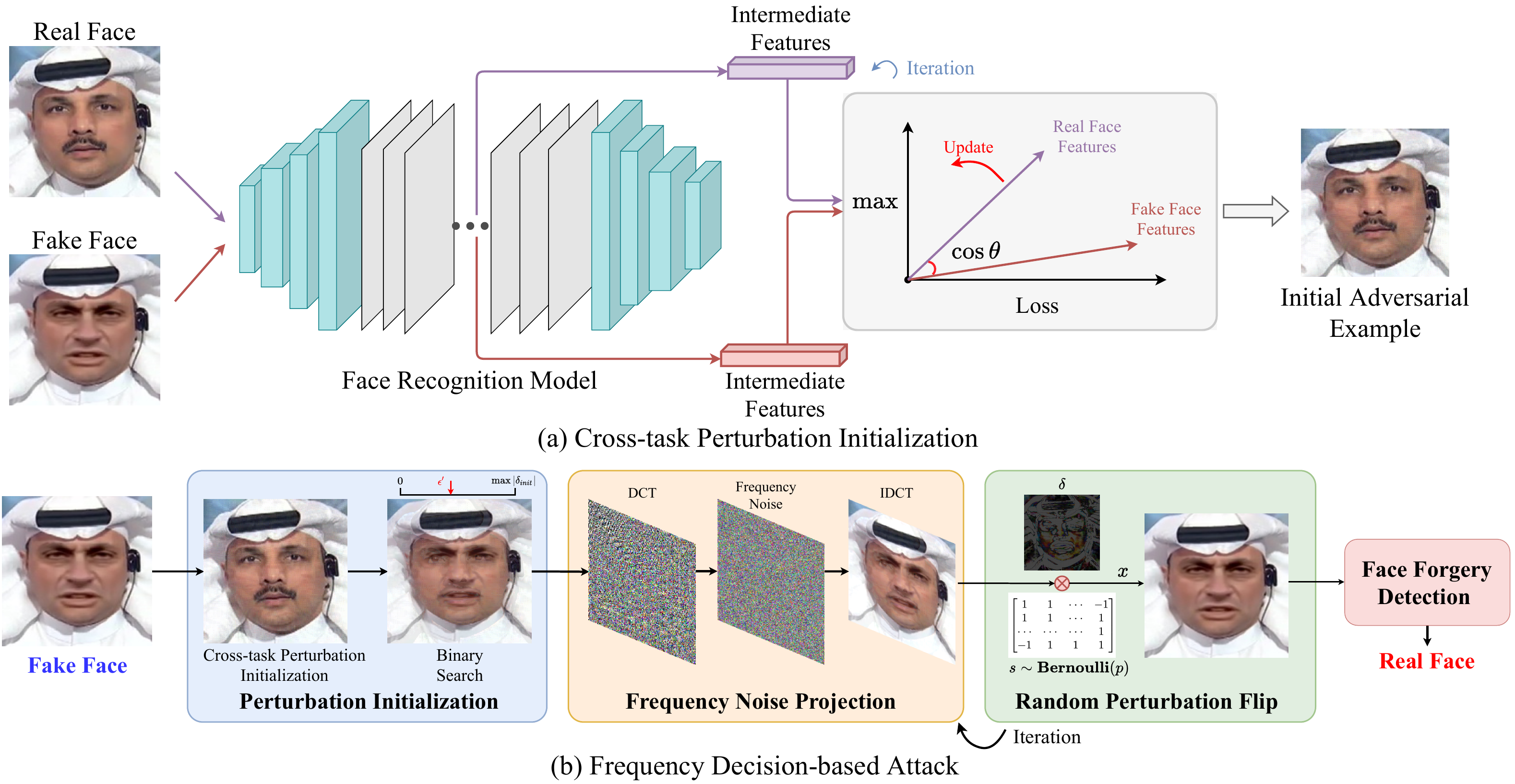}
    \end{center}
    \caption{(a) The overview of cross-task perturbation initialization. With the high correlation of intermediate features between face-related tasks, we iteratively update the real face, keeping it adversarial while improving efficiency for subsequent attacks. (b) The pipeline of the frequency decision-based attack. We first obtain the initial perturbation by perturbation initialization. Then, we iterate in frequency noise projection and random perturbation flip until $c(x_{adv})=0$ and $||\delta||_\infty \leq \epsilon$.}
    \label{fig:cpi}
\end{figure*}

\subsection{Cross-task Perturbation Initialization}
\label{sec:cpi}
In this section, we introduce the cross-task perturbation initialization (CPI) to handle the attack initialization failure. Actually, the easiest way to explore decision-based attacks in face forgery detection is to apply existing decision-based attacks directly~\cite{boundary,OptAttack,signopt, hsja,SignFlip,TA}. In the case study, we choose the state-of-the-art HSJA~\cite{hsja}, SignFlip~\cite{SignFlip}, and RayS~\cite{RayS}. We apply these attacks directly to face forgery detection and find that they cause the attack to fail. Specifically, they all have a situation where the initialization fails, and the attack success rate is close to 0\%. 
It shows that the random initialization is unsuccessful and cannot guarantee adversarial, as shown in the random initialization in Figure~\ref{fig:bypass}. 
Face forgery detection usually recognizes faces with large random noises as fake faces because adding large random noises is a common manipulation.
In addition, due to the large random noise, such samples usually fail face recognition.
Furthermore, using real faces as initialization can guarantee adversarial, but it may lead to artifacts, so image quality is not guaranteed. Hence, we hope a perturbation initialization can handle this issue while ensuring query efficiency.

Naturally, exploiting the adversarial transferability of face forgery detectors is a possible strategy. However, in practice, we not only do not have access to the training data of the threat model but also do not know the face forgery generation method it adopts, which greatly limits the possibility of using transferable adversarial perturbations. 
Since face recognition is a very common face-related task and its open-source model is relatively easy to obtain, it may help the attack achieve perturbation initialization. 
Furthermore, intermediate features may have high correlations between different face-related tasks. From this perspective, using face recognition models to transfer adversarial perturbations is a feasible initialization strategy.

\begin{algorithm}[t]
  \caption{Cross-task Perturbation Initialization (CPI)}
  \label{alg:cpi}
  \textbf{Input}: Fake face $x$, real face $x_r$, face recognition model $Z$ \\ \textbf{Parameters}: Maximum perturbation $\xi$, layer $l$, iteration number $K$
  \\
  \textbf{Output}: Adversarial face $x_{adv}$
  \begin{algorithmic}[1]
  \STATE $x_0 \leftarrow x_r$, $\delta_0 \leftarrow 0$
  \FOR{$i \in [1,\ K]$}
    \STATE $J(x_{i-1},x) \leftarrow \mathrm{-Cos}(Z_l(x_{i-1}),\ Z_l(x))$
    \STATE $\delta_{i} \leftarrow \delta_{i-1} + \frac{\xi}{K} \cdot  \mathrm{sign}(\nabla_{x_{i-1}}J(x_{i-1},x))$
    \STATE $x_i \leftarrow \mathrm{Clip}_{x_r,\ \xi}(x_{i-1}+\delta_i)$
  \ENDFOR
  \STATE $x_{adv} \leftarrow x_{K}$
    \RETURN $x_{adv}$
  \end{algorithmic}
\end{algorithm}

We assume that if there is a high degree of similarity between the intermediate features of the face recognition model and face forgery models, then attacking the face recognition model is equivalent to attacking the face forgery detection models.
To verify the correlation of intermediate features across face tasks, we conduct an empirical study on FaceForensics++~\cite{ffpp}. We randomly select 256 frames to calculate the intermediate features and then average for calculating cosine similarity. Figure~\ref{fig:feature} shows the cosine similarity of intermediate features between face forgery detectors and face recognition models. Here, the face recognition models are FaceNet~\cite{FaceNet}, CosFace~\cite{CosFace}, and ArcFace~\cite{ArcFace}. The face forgery detection models are ResNet50~\cite{resnet}, Xception~\cite{xception}, EfficientNet-b4 (Eb4)~\cite{EfficientNet}. The middle layer selection of face recognition models is shown in Table~\ref{tab:layerselection} and the face forgery detectors select the second block unit. According to Figure~\ref{fig:feature}, we find that face recognition models and face forgery detectors have a high correlation between the intermediate features of the same face because these are all face-related tasks and models capture the general features of faces.

The above observations illustrate the high correlation of intermediate features between face-related tasks. Due to the high correlation between the intermediate features of face recognition and face forgery detection, the perturbation of attacking face recognition can also perturb face forgery detection to a certain extent, so we propose cross-task perturbation initialization. Given a face recognition model $Z$, $l$ represents the $l$-th layer, and $Z_l(\cdot)$ represents the intermediate feature of the $l$-th layer. Considering attacking face forgery detectors, the adversary has a real face $x_r$ and a forged fake face $x$. Usually, $x_r$ is selected as the initialization to ensure adversarially, and then the distance from $x$ is reduced. Therefore, the optimization objective for cross-task adversarial perturbation is as follows:
\begin{equation}
    \mathop{\mathrm{argmin}}\limits_{\delta}\ \mathrm{Cos}(Z_l(x),\ Z_l(x_r+\delta)), \quad s.t. \quad ||\delta||_\infty \leq \xi,
\end{equation}
where $\xi$ is maximum cross-task perturbation and $\mathrm{Cos}$ calculates the cosine similarity between $Z_l(x)$ and $Z_l(x_r+\delta)$. 
By minimizing the cosine similarity between $Z_l(x)$ and $Z_l(x_r+\delta)$, we keep $x_r+\delta$ away from fake faces $x$ in the feature space and move it towards real faces, thus providing a more correct attack direction and improve query efficiency.

Algorithm~\ref{alg:cpi} illustrates the whole pipeline of the cross-task perturbation initialization (CPI), where $\mathrm{sign}$ represents the sign operation on the gradient and $\mathrm{Clip}_{x_r,\ \xi}(\cdot)$ projects the perturbation $\delta$ to $\xi$-ball in order to satisfy $||\delta||_\infty \leq \xi$. 
The time complexity of CPI is $O(K)$, so the calculation of CPI requires $K$ times of forward and backward propagation of the face recognition model. 
In this paper, $K=10$. We test the time to calculate CPI on a NVIDIA A100, which is about $50.34$ms for an image, with almost no additional overhead.

To sum up, we leverage the fast sign gradient to generate perturbations efficiently and focus on the high feature correlation between face-related tasks, which is beneficial to the generalizability of our method to adversarial attacks and defenses on more face-related tasks. As facial recognition models constitute fundamental infrastructure for face-related tasks, they extract the most essential facial features. Face forgery detection, being another face-centric task, is also highly sensitive to these features. If the high correlation between intermediate features of facial recognition and face forgery detection, perturbations designed to attack facial recognition models can consequently disrupt face forgery detection to a significant extent. As convincingly revealed in Figure 2 of the main body, the intermediate features of facial recognition models and face forgery detectors exhibit strong correlation when processing identical face images. Therefore, CPI provides superior initialization for attacking face forgery detection systems. This ensures a more precise optimization trajectory while enabling simultaneous attacks on multiple face-related intelligent systems.

\subsection{Frequency Decision-based Attack}
Recent work~\cite{fre_analysis,f3net,M2TR} on face forgery detection has illustrated a significant difference in frequency between real and fake faces. In addition, some work~\cite{highfre,Fourier,SSA} also shows that different models rely on different frequency components to make decisions. Inspired by these works, we consider adding noise directly to the frequency of human faces, \textit{aiming to eliminate the difference between real and fake faces in frequency to improve the attack efficiency.}

\begin{algorithm}[t]
  \caption{Binary Search}
  \label{alg:binarysearch}
  \textbf{Input}: Fake face $x$, initial perturbation $\delta_{init}$\\ \textbf{Parameters}:  iteration number $k$\\
  \textbf{Output}: Final initial perturbation $\delta$ 
  
  \begin{algorithmic}[1]
  \STATE $l \leftarrow 0$, $u \leftarrow \max(|\delta_{init}|)$
  \FOR{$i \in [1,\ k]$}
    \STATE $m \leftarrow (l+u)/2$
    \STATE $x_t \leftarrow \mathrm{Clip}_{x_r,\ m}(x+\delta_{init})$
    \IF{$c(x_t)=0$}
        \STATE $u \leftarrow m$
    \ELSE
        \STATE $l \leftarrow m$
    \ENDIF
  \ENDFOR
  \STATE $\delta \leftarrow \mathrm{Clip}_{0,\ u}(\delta_{init})$
    \RETURN $\delta$
  \end{algorithmic}
\end{algorithm}

Based on the general framework of decision-based attacks~\cite{hsja,SignFlip}, we propose the frequency decision-based attack (FDA) on face forgery detection. Figure~\ref{fig:cpi} summarizes the pipeline of our proposed attack, including the perturbation initialization, the frequency noise projection, and the random perturbation flip. Specifically, we first generate the cross-task adversarial perturbation $\delta_{init}$ by perturbation initialization and resort to binary search to reduce the perturbation magnitude and guarantee adversarial. In frequency noise projection, we add perturbations in the frequency domain and constrain image quality in the spatial domain. Then, we randomly inverse the sign of the partial perturbation, expecting a better adversarial perturbation. Finally, we iterate in frequency noise projection and random perturbation flip until $c(x_{adv})=0$ and $||\delta||_\infty \leq \epsilon$.

\textbf{Perturbation Initialization.}
Generally speaking, the perturbation initialization is used in decision-based attacks to make clean examples keep adversarial. Given a face classified as real, we utilize the cross-task perturbation initialization to obtain initial perturbations $\delta_{init}$, which satisfies the adversarial constraint. Note that cross-task adversarial perturbations are imperceptible, so faces after cross-task perturbation initialization remain adversarial. Further, in order to reduce the magnitude of the initial perturbation, we choose to apply the binary search to make the initial adversarial face as close to $x$ as possible. Binary search in perturbation initialization is summarized in Algorithm~\ref{alg:binarysearch}. Given an initial perturbation $\delta_{init}$, we continuously dichotomize the perturbation range from $[0, \max(\delta_{init})]$ on the premise of ensuring adversarial, and finally obtain the final adversarial perturbation $\delta$. In this paper, we set $k$ to 10. With the help of binary search, we obtain the final initial perturbation $\delta$, where $||\delta||_\infty \leq \epsilon'$ and $ \epsilon' \in (0, \max|\delta_{init}|]$.

\begin{algorithm}[t]
  \caption{Frequency Decision-based Attack}
  \label{alg:fda}
  \textbf{Input}: Fake face $x$, real face $x_r$, face recognition model $Z$, forgery detector $f$
  \\ 
  \textbf{Parameters}: Maximum perturbation $\epsilon$, step size $\gamma$
  \\
  \textbf{Output}: Adversarial face $x_{adv}$
  \begin{algorithmic}[1]
  \STATE $\mathrm{\#\ Perturbation\ Initialization}$
  \STATE {$\delta_{init} \leftarrow \mathrm{\textbf{CPI}}(x, x_r, Z)$ }
  \STATE $\delta \leftarrow \mathrm{\textbf{BinarySearch}}(x, \delta_{init})$
  \STATE $\epsilon' \leftarrow ||\delta||_\infty$
  \WHILE{$\epsilon < \epsilon'$}
    \STATE $\mathrm{\#\ Frequency\ Noise\ Projection}$
    \STATE $x_{dct}^p \leftarrow \mathcal{D}(x+\delta)$ 
    \STATE $\eta \sim \{-\gamma, \gamma \}^d$
    \STATE $x_{idct}^p \leftarrow \mathcal{D'}(x_{dct}^p + \eta)$
    \STATE $x^p \leftarrow \mathrm{Clip}_{x,\ \epsilon'-\kappa}(x_{dct}^p)$
    \IF{$c(x^p)=0$}
        \STATE $\delta \leftarrow x^p - x$
    \ENDIF
    \STATE $\mathrm{\#\ Perturbation\ Random\ Flip}$
    \STATE $s \sim \mathrm{\textbf{Bernoulli}}(p)$ 
    \STATE $\delta^p \leftarrow \delta \odot s$
    \IF{$c(x+\delta^p)==0$}
        \STATE $\delta \leftarrow \delta^p$
    \ENDIF
    \STATE $\epsilon' \leftarrow ||\delta||_\infty$
  \ENDWHILE
  \STATE $x_{adv} \leftarrow x + \delta$
    \RETURN $x_{adv}$
  \end{algorithmic}
\end{algorithm}

\textbf{Frequency Noise Projection.}
In frequency noise projection, we add perturbations in the frequency domain and constrain image quality in the spatial domain. Here, $\mathcal{D}(\cdot)$ denotes discrete cosine transform (DCT) and $\mathcal{D'}(\cdot)$ denotes inverse discrete cosine transform (IDCT). First, we transform the image from the spatial domain to the frequency domain with DCT. Then, we add a random noise in the frequency domain, where $\eta \sim \{-\gamma, \gamma \}^d$. Finally, we project it to the $\epsilon'$-ball in the spatial domain and gradually decrease $\epsilon'$ until it is satisfied $\epsilon' \leq \epsilon$. To sum up, the specific calculation is as follows:
\begin{equation}
    \label{fre_noise}
    x_{idct} \leftarrow \mathcal{D'}(\mathcal{D}(x_{adv}) + \eta),\quad \eta \sim \{-\gamma, \gamma \}^d, 
\end{equation}
\vspace{-24pt}
\begin{equation}
    x_{adv} \leftarrow \mathrm{Clip}_{x,\ \epsilon' -\kappa}(x_{idct}),
\end{equation}
where $\kappa$ is 0.004. For convenience, we perform DCT and IDCT on the whole image. Note that $x$ and $\mathcal{D'}(\mathcal{D}(x))$ are equivalent. However, after adding the noise $\eta$ to $\mathcal{D}(x)$ on the frequency domain, $x$ and $\mathcal{D'}(\mathcal{D}(x) + \eta)$ are not equivalent. $\eta$ is transformed into a frequency domain distribution with the help of frequency information according to IDCT.

\textbf{Random Perturbation Flip.}
Existing work~\cite{PriorConvictions,Parsimonious} shows that searching on the nodes of the $l_\infty$ ball is more efficient than searching in the $l_\infty$ ball. Motivated by these observations, both~\cite{SignFlip} and~\cite{RayS} suggest some strategies based on searching on the $l_\infty$ ball. Here, we choose~\cite{SignFlip} to further accelerate the attack.
After the frequency noise projection, we randomly partial coordinates to flip the signs of perturbation. Suppose $s \in \{-1,1\}^d$ and $p \in (0, 1)^d$, the random perturbation flip is formulated by:
\begin{equation}
    s \sim \mathrm{\textbf{Bernoulli}}(p), \quad \delta_s \leftarrow \delta_p \odot s.
\end{equation}
The frequency decision-based attack is shown in Algorithm~\ref{alg:fda} and its time complexity is $O(N)$, where $N$ is the maximum query budget. The main time cost of FDA is the communication and inference time with the API (or offline model). In addition, the calculation of Frequency Noise Projection and Perturbation Random Flip has almost no additional computational cost.
Since $N>>K$, the total time complexity is $O(N)$.
To sum up, based on the cross-task perturbation initialization, the frequency decision-based attack can threaten face forgery detection and better ensure the image quality, which is convenient for the application of downstream tasks.

\begin{table}[t]
\caption{The accuracy of spatial-based and frequency-based face forgery detectors on the CelebDF and FFDF datasets.}
\label{tab:acc}
\setlength{\tabcolsep}{2.mm}
\centering
\scalebox{0.83}{
\begin{tabular}{@{}c|cccc|cc@{}}
\toprule
Model & ResNet50 & Xception & Eb4 & ViT & F$^3$Net & M2TR \\ \midrule
CelebDF & 98.51 & 99.05 & 99.44 & 96.73 & 96.47 & 99.76 \\
FFDF & 94.87 & 95.24 & 95.61 & 93.45 & 97.52 & 97.93 \\
\bottomrule
\end{tabular}
}
\end{table}

\section{Experiments}
\label{sec:experiments}
In this section, we verify the effectiveness of the proposed method. First, we introduce the relevant settings of the experiments. We then provide experimental results on state-of-the-art detectors to illustrate the performance of our attack. Next, we conduct the ablation study to evaluate the effectiveness of the proposed module. Further, we evaluate the image quality of the generated adversarial examples. Finally, we further demonstrate the effectiveness of our method on industrial APIs, adversarial transferability, and adversarial defenses.

\subsection{Experimental Setup}
\textbf{Datasets.} 
FaceForensics++ (FFDF)~\cite{ffpp} is a popular large-scale face forgery dataset containing 1,000 video sequences. It includes four forging methods, Face2Face~\cite{Face2Face}, FaceSwap~\cite{Faceswap}, DeepFakes~\cite{Deepfake} and NeuralTextures~\cite{neuraltextures}. We randomly select 4 frames of each video in the test set, a total of 560 frames (140$\times$4=560). CelebDF~\cite{CelebDF} is a large-scale DeepFake video dataset with a total of 5,639 DeepFake videos. We also randomly select 500 frames from different videos from the testset.

\textbf{Models.}
Face forgery detection is based on the spatial domain or frequency domain. For spatial-based face forgery detection, we choose four spatial-based classification networks, such as ResNet50~\cite{resnet}, Xception~\cite{xception}, EfficientNet-b4 (Eb4)~\cite{EfficientNet} and ViT-B (ViT)~\cite{vit}. For frequency-based face forgery detection, we choose state-of-the-art models, such as F$^3$Net~\cite{f3net} and M2TR~\cite{M2TR}. 
All the models are trained according to the corresponding paper settings. The accuracy of these face forgery detectors is shown in Table~\ref{tab:acc}. 
For face recognition models, we choose FaceNet~\cite{FaceNet} to initialize cross-task perturbations. The weights and implementation of FaceNet\footnote{\url{https://github.com/ShawnXYang/Face-Robustness-Benchmark/blob/master/RobFR/networks/FaceNet.py}} are based on Face-Robustness-Benchmark (RobFR)\footnote{\url{https://github.com/ShawnXYang/Face-Robustness-Benchmark}}.

\begin{table*}[t]
\caption{Attack results of decision-based attacks against spatial-based face forgery detection on FaceForensics++ and CelebDF.}
\label{tab:spatial}
\centering
\setlength{\tabcolsep}{1.6mm}
\scalebox{0.82}{
\begin{tabular}{@{}c|c|ccc|ccc|ccc|ccc@{}}
\toprule
\multirow{2}{*}{Dataset} & Model & \multicolumn{3}{c|}{ResNet50} & \multicolumn{3}{c|}{Xception} & \multicolumn{3}{c|}{Eb4} & \multicolumn{3}{c}{ViT} \\ \cmidrule(l){2-14} 
 & Method & AQ$\downarrow$ & MQ$\downarrow$ & ASR(\%)$\uparrow$ & AQ$\downarrow$ & MQ$\downarrow$ & ASR(\%)$\uparrow$ & AQ$\downarrow$ & MQ$\downarrow$ & ASR(\%)$\uparrow$ & AQ$\downarrow$ & MQ$\downarrow$ & ASR(\%)$\uparrow$ \\ \midrule
\multirow{8}{*}{FFDF} & Boundary & 6607.10 & 6363.0 & \multicolumn{1}{c|}{5.50} & 5165.22 & 4819.0 & \multicolumn{1}{c|}{15.20} & - & - & \multicolumn{1}{c|}{0.00} & 3679.73 & 3473.0 & 8.04 \\
 & Opt & 5105.91 & 4883.0 & \multicolumn{1}{c|}{65.17} & 4490.10 & 4204.0 & \multicolumn{1}{c|}{82.14} & 4826.47 & 4394.0 & \multicolumn{1}{c|}{69.11} & 3393.52 & 3482.0 & 30.00 \\
 & SignOpt & 1780.87 & 1552.0 & \multicolumn{1}{c|}{5.54} & 4113.86 & 3328.0 & \multicolumn{1}{c|}{20.11} & 1342.23 & 831.0 & \multicolumn{1}{c|}{11.07} & 1342.22 & 831.0 & 11.17 \\
 & TA & 1907.89 & 981.0 & \multicolumn{1}{c|}{74.11} & 1156.51 & 235.0 & \multicolumn{1}{c|}{61.90} & 983.97 & 174.0 & \multicolumn{1}{c|}{49.46} & 529.64 & 67.0 & 91.07 \\
 & HSJA & 198.02 & 190.0 & \multicolumn{1}{c|}{\textbf{100.00}} & 186.55 & 186.0 & \multicolumn{1}{c|}{\textbf{100.00}} & 246.25 & 191.0 & \multicolumn{1}{c|}{96.25} & 245.19 & 188.0 & 79.64 \\
 & SignFlip & 53.51 & 50.0 & \multicolumn{1}{c|}{\textbf{100.00}} & 63.93 & 50.0 & \multicolumn{1}{c|}{\textbf{100.00}} & - & - & \multicolumn{1}{c|}{0.00} & 167.85 & 78.0 & 99.28 \\
 & RayS & 152.84 & 133.0 & \multicolumn{1}{c|}{\textbf{100.00}} & 1965.08 & 1580.0 & \multicolumn{1}{c|}{\textbf{100.00}} & - & - & \multicolumn{1}{c|}{0.00} & 312.35 & 270.0 & \textbf{100.00} \\
 & Ours & \textbf{36.16} & \textbf{30.0} & \multicolumn{1}{c|}{\textbf{100.00}} & \textbf{12.72} & \textbf{12.0} & \multicolumn{1}{c|}{\textbf{100.00}} & \textbf{39.94} & \textbf{34.0} & \multicolumn{1}{c|}{\textbf{96.96}} & \textbf{139.86} & \textbf{42.0} & \textbf{100.00} \\ \midrule
\multirow{8}{*}{CelebDF} & Boundary & - & - & \multicolumn{1}{c|}{0.00} & 4159.81 & 3594.0 & \multicolumn{1}{c|}{10.40} & 4438.24 & 3349.0 & \multicolumn{1}{c|}{5.00} & 5586.00 & 5460.0 & 0.60 \\
 & Opt & 6939.37 & 7374.0 & \multicolumn{1}{c|}{20.40} & 2101.52 & 1340.0 & \multicolumn{1}{c|}{96.00} & 3925.10 & 3354.0 & \multicolumn{1}{c|}{87.60} & 5800.34 & 6031.0 & 18.60 \\
 & SignOpt & 3541.41 & 3124.0 & \multicolumn{1}{c|}{37.80} & 1260.02 & 926.0 & \multicolumn{1}{c|}{73.00} & 1752.79 & 1737.0 & \multicolumn{1}{c|}{60.8} & 3282.36 & 2873.0 & 25.00 \\
 & TA & 171.55 & 116.0 & \multicolumn{1}{c|}{90.60} & 194.39 & 126.0 & \multicolumn{1}{c|}{86.80} & 75.92 & 31.0 & \multicolumn{1}{c|}{97.6} & 238.82 & 178.0 & 78.60 \\
 & HSJA & 232.74 & 189.0 & \multicolumn{1}{c|}{70.00} & 187.11 & 186.0 & \multicolumn{1}{c|}{99.20} & 188.24 & 188.0 & \multicolumn{1}{c|}{\textbf{100.00}} & 1595.12 & 327.0 & 18.00 \\
 & SignFlip & 33.39 & \textbf{12.0} & \multicolumn{1}{c|}{99.60} & 15.25 & \textbf{12.0} & \multicolumn{1}{c|}{99.20} & 20.32 & \textbf{12.0} & \multicolumn{1}{c|}{\textbf{100.00}} & 288.99 & 98.0 & 99.60 \\
 & RayS & - & - & \multicolumn{1}{c|}{0.00} & 203.16 & 175.0 & \multicolumn{1}{c|}{\textbf{100.00}} & 169.85 & 140.0 & \multicolumn{1}{c|}{\textbf{100.00}} & \textbf{237.76} & 215.0 & \textbf{100.00} \\
 & Ours & \textbf{32.98} & \textbf{12.0} & \multicolumn{1}{c|}{\textbf{100.00}} & \textbf{12.52} & \textbf{12.0} & \multicolumn{1}{c|}{\textbf{100.00}} & \textbf{15.48} & \textbf{12.0} & \multicolumn{1}{c|}{\textbf{100.00}} & 283.34 & \textbf{92.0} & \textbf{100.00} \\ \bottomrule
\end{tabular}
}
\end{table*}

\begin{table*}[t]
\caption{Attack results of decision-based attacks against frequency-based detectors on FaceForensics++ and CelebDF.}
\label{tab:frequency}
\setlength{\tabcolsep}{1.2mm}
\centering
\scalebox{0.8}{
\begin{tabular}{@{}c|c|ccc|ccc|c|ccc|ccc@{}}
\toprule
Model &  & \multicolumn{3}{c|}{F$^3$Net} & \multicolumn{3}{c|}{M2TR} &  & \multicolumn{3}{c|}{F$^3$Net} & \multicolumn{3}{c}{M2TR} \\ \cmidrule(r){1-1} \cmidrule(lr){3-8} \cmidrule(l){10-15} 
Method & \multirow{-2}{*}{Dataset} & AQ$\downarrow$ & MQ$\downarrow$ & ASR (\%)$\uparrow$ & AQ$\downarrow$ & MQ$\downarrow$ & ASR (\%)$\uparrow$ & \multirow{-2}{*}{Dataset} & AQ$\downarrow$ & MQ$\downarrow$ & ASR (\%)$\uparrow$ & AQ$\downarrow$ & MQ$\downarrow$ & ASR (\%)$\uparrow$ \\ \midrule
Boundary &  & 3313.41 & 2456.0 & 12.06 & 2202.50 & 1427.0 & 3.04 &  & 3393.24 & 2883.0 & 22.00 & 1920.33 & 14.0 & 1.80 \\
Opt &  & 3435.38 & 3286.0 & 5.53 & 3549.99 & 3029.0 & 79.46 &  & 1839.61 & 1178.0 & 95.60 & 5053.90 & 5363.0 & 30.60 \\
SignOpt &  & 4262.19 & 3956.0 & 19.56 & 1078.89 & 785.0 & 45.90 &  & 1220.62 & 868.0 & 59.20 & 2752.79 & 2549.0 & 71.20 \\
TA &  & {\color[HTML]{000000} 1143.31} & {\color[HTML]{000000} 293.0} & {\color[HTML]{000000} 50.53} & {\color[HTML]{000000} 837.70} & {\color[HTML]{000000} 305.0} & {\color[HTML]{000000} 58.75} &  & {\color[HTML]{000000} 179.04} & {\color[HTML]{000000} 97.0} & {\color[HTML]{000000} 88.00} & {\color[HTML]{000000} 220.31} & {\color[HTML]{000000} 162.0} & {\color[HTML]{000000} 87.40} \\
HSJA &  & 199.38 & 187.0 & 82.50 & 886.46 & 188.0 & 81.42 &  & 187.04 & 186.0 & \textbf{100.00} & 1275.91 & 324.0 & 33.60 \\
SignFlip &  & 732.11 & 280.0 & 93.57 & 102.07 & 86.0 & \textbf{100.00} &  & 19.54 & \textbf{12.0} & 97.20 & 103.84 & \textbf{12.0} & \textbf{100.00} \\
RayS &  & \textbf{140.30} & 117.0 & 99.92 & 225.33 & 157.0 & \textbf{100.00} &  & 84.54 & 77.0 & \textbf{100.00} & 84.21 & 79.0 & \textbf{100.00} \\
Ours & \multirow{-8}{*}{FFDF} & 152.25 & \textbf{114.0} & \textbf{100.00} & \textbf{79.19} & \textbf{60.0} & \textbf{100.00} & \multirow{-8}{*}{CelebDF} & \textbf{12.59} & \textbf{12.0} & \textbf{100.00} & \textbf{32.67} & \textbf{12.0} & \textbf{100.00} \\ \bottomrule
\end{tabular}
}
\end{table*}

\textbf{Attack Methods.}
We compare the proposed algorithm with state-of-the-art attack algorithms, including Boundary Attack~\cite{boundary}, Opt-Attack~\cite{OptAttack}, Sign-Opt~\cite{signopt}, Triangle Attack (TA)~\cite{TA}, HSJA~\cite{hsja}, SignFilp~\cite{SignFlip}, and RayS~\cite{RayS}. Attack parameters are consistent with corresponding attacks.

\textbf{Evaluation metrics.}
We choose three metrics to quantitatively evaluate the performance of attack methods: attack success rate (ASR), average queries (AQ), and median queries (MQ). Attack success rate (ASR) is defined as the proportion of faces that are successfully attacked. Here, a successful attack is to classify fake faces as real faces. Average queries (AQ) and median queries (MQ) are only calculated over successful attacks, following~\cite{SignFlip} and~\cite{RayS}.

\textbf{Implementation details.} 
We use Retinaface~\cite{retinaface} to detect and crop faces and follow the open-source PyDeepFakeDet\footnote{\url{https://github.com/wangjk666/PyDeepFakeDet/blob/main/DATASET.md}} for preprocessing. For the cropped face, the model resizes it to an image shape and then uses it as the model's input. 
For decision-based attacks, we set the maximum perturbation of each pixel to be $\epsilon=0.05$ and the maximum query budget is 10,000. Since these attacks are initialized with randomness, it is easy to cause the attack to fail, so we choose to use the same real face as the initialization if the attack allows. For cross-task perturbation initialization, the cosine similarity of intermediate layer features is interpolated to the same dimension. Here, $K=10$, $\gamma=1.75$, $\xi=0.031$, and $p=0.999$. In the frequency decision-based attack, the attack ends when $\epsilon \geq \epsilon'$ or the query reaches the maximum number of queries.

\begin{table*}[t]
\caption{Attack performance with different maximum perturbation $\epsilon$ on FaceForensics++.}
\label{tab:attackeps}
\scalebox{0.9}{
\begin{tabular}{@{}c|c|ccc|ccc|ccc@{}}
\toprule
\multirow{2}{*}{Model} & Model & \multicolumn{3}{c|}{$\epsilon=0.031$} & \multicolumn{3}{c|}{$\epsilon=0.05$} & \multicolumn{3}{c}{$\epsilon=0.062$} \\ \cmidrule(l){2-11} 
 & Method & AQ$\downarrow$ & MQ$\downarrow$ & ASR (\%)$\uparrow$ & AQ$\downarrow$ & MQ$\downarrow$ & ASR (\%)$\uparrow$ & AQ$\downarrow$ & MQ$\downarrow$ & ASR (\%)$\uparrow$ \\ \midrule
\multirow{8}{*}{ResNet50} & Boundary & 4326.10 & 3826.0 & 1.96 & 6607.10 & 6363.0 & 5.50 & 5224.28 & 4774.0 & 5.89 \\
 & Opt & 5848.47 & 5615.0 & 22.19 & 5105.91 & 4883.0 & 65.17 & 4343.85 & 4026.0 & 83.93 \\
 & SignOpt & 3065.18 & 2893.0 & 6.07 & 1780.87 & 1552.0 & 5.54 & 1508.39 & 1133.0 & 6.07 \\
 & TA & 2050.71 & 1289.0 & 57.32 & 1907.89 & 981.0 & 74.11 & 1829.67 & 918.0 & 79.46 \\
 & HSJA & 220.34 & 190.0 & 91.07 & 198.02 & 190.0 & \textbf{100.00} & 195.64 & 190.0 & \textbf{100.00} \\
 & SignFlip & 105.58 & 98.0 & \textbf{100.00} & 53.51 & 50.0 & \textbf{100.00} & 48.75 & 40.0 & \textbf{100.00} \\
 & RayS & 213.64 & 112.0 & \textbf{100.00} & 152.84 & 133.0 & \textbf{100.00} & 141.78 & 126.0 & \textbf{100.00} \\
 & Ours & \textbf{101.38} & \textbf{76.0} & \textbf{100.00} & \textbf{36.16} & \textbf{30.0} & \textbf{100.00} & \textbf{32.29} & \textbf{22.0} & \textbf{100.00} \\ \midrule
\multirow{8}{*}{ViT-B} & Boundary & 3328.29 & 2740.0 & 5.35 & 3679.73 & 3473.0 & 8.04 & 3118.15 & 2890.0 & 8.39 \\
 & Opt & 5130.87 & 4622.0 & 18.51 & 3393.52 & 3482.0 & 30.00 & 4222.17 & 3753.0 & 64.28 \\
 & SignOpt & 2451.80 & 1872.0 & 11.17 & 1342.22 & 831.0 & 11.17 & 1176.37 & 787.0 & 11.17 \\
 & TA & 835.56 & \textbf{172.0} & 72.14 & 529.64 & 67.0 & 91.07 & 344.77 & 46.0 & 95.71 \\
 & HSJA & \textbf{288.08} & 188.0 & 38.03 & 245.19 & 188.0 & 79.64 & 264.23 & 188.0 & 91.61 \\
 & SignFlip & 644.30 & 244.0 & 98.03 & 167.85 & 78.0 & 99.28 & 105.81 & 64.0 & 99.10 \\
 & RayS & 678.24 & 367.0 & \textbf{100.00} & 312.35 & 270.0 & \textbf{100.00} & 273.83 & 243.0 & \textbf{100.00} \\
 & Ours & 600.15 & 330.0 & \textbf{100.00} & \textbf{139.86} & \textbf{42.0} & \textbf{100.00} & \textbf{86.58} & \textbf{42.0} & \textbf{100.00} \\ \bottomrule
\end{tabular}
}
\end{table*}

\begin{table}[t]
\caption{Attack performance of SBI on FaceForensics++.}
\centering
\label{tab:sbi}
\setlength{\tabcolsep}{4mm}
\scalebox{0.83}{
\begin{tabular}{@{}c|ccc@{}}
\toprule
Model & \multicolumn{3}{c}{SBI} \\ \midrule
Method & AQ$\downarrow$ & MQ$\downarrow$ & ASR (\%)$\uparrow$ \\ \midrule
Boundary & - & - & 0 \\
Opt & - & - & 0 \\
SignOpt & - & - & 0 \\
TA & - & - & 0 \\
HSJA & - & - & 0 \\
SignFlip & - & - & 0 \\
Rays & 544.32 & 402.0 & \textbf{100.00} \\
Ours & \textbf{154.23} & \textbf{60.0} & \textbf{100.00} \\ \bottomrule
\end{tabular}
}
\end{table}

\subsection{Attacks on State-of-the-art Detectors}
\textbf{Attack on Spatial-based Models.}
Table~\ref{tab:spatial} shows the attack success rates against spatial-based models on FaceForensics++ and CelebDF. Compared with other attacks, our method has a higher query efficiency on the premise of ensuring 100\% attack success rate. Furthermore, Boundary, SignFlip, and RayS fail to attack some models (e.g. ResNet50, Eb4) even with real faces as perturbation initialization. For example, SignFlip and RayS attack Eb4 fails on FaceForensics++, but our method achieves the attack with a high attack efficiency. Meanwhile, compared to SignFlip and RayS, the average query number of our method is about 1/2 and 1/4 of that on ResNet50, and 1/5 and 1/160 of that on Xception. On CelebDF, our method improves the attack efficiency by 17.90\% and 23.81\% against Xception and Eb4 compared to the state-of-the-art SignFlip. Although less related to the features of ViT, our method can also achieve a state-of-the-art attack success rate, which effectively illustrates the generalization of our method.

\textbf{Attack on Frequency-based Models.}
Face forgery detection often uses the difference in frequency as a clue to judge whether it is forgery. To verify the adversarial robustness of these frequency-based methods, we choose F$^3$Net~\cite{f3net} and M2TR~\cite{M2TR} for experiments. Table~\ref{tab:frequency} shows the attack success rates of frequency-based detectors on FaceForensics++ and CelebDF. Frequency-based detectors perform well on face forgery detection, but they are just as susceptible to adversarial robustness as spatial-based detectors. On F$^3$Net, our method achieves state-of-the-art attack performance, especially on CelebDF, with 35.56\% fewer queries on average than SignFlip. On M2TR, our method improves the query efficiency by 22.41\% and 61.20\% over state-of-the-art methods on FaceForensics++ and CelebDF, respectively. Although frequency-based face forgery detectors have better detection performance than spatial-based detectors, they are both vulnerable to adversarial examples. Our attack adds noise in the frequency domain and has high efficiency in attacking both spatial and frequency domain based detectors, which illustrates the effectiveness of frequency noise. Therefore, the above experiments demonstrate the effectiveness of our method and further expose the adversarial vulnerability of face forgery detection.

\textbf{Attack on Self-blended Models.} 
\cite{sbi} presents an innovative deepfake detection method utilizing self-blended images (SBIs), which are artificially generated by blending pseudo source and target images derived from a single authentic image. This approach replicates typical forgery artifacts, such as blending boundaries and statistical discrepancies, thus producing more generalized and less distinguishable fake samples. Table~\ref{tab:sbi} shows the attack performance of SBI\footnote{We use the implementation and weights (FF-c23) provided by \url{https://github.com/mapooon/SelfBlendedImages}.} on FaceForensics++~\cite{ffpp}. Except for Rays and Ours, other methods use target images as initialization. For SBI, these are obviously blended images, so they will always be identified as fake faces, resulting in unsuccessful attacks. Due to the efficient initialization provided by CPI, our attack maintains a 100\% attack success rate (ASR) on SBI, while requiring only approximately 28.33\% of the average queries needed by RayS. This demonstrates the continued effectiveness of our method, even against more generalized face forgery detectors.

\textbf{Attack on different perturbations $\epsilon$.} To further improve our adversarial robustness evaluation, we test the attack performance with different maximum perturbations $\epsilon$ on FaceForensics++’s ResNet and ViT-B, as shown in Table~\ref{tab:attackeps}. We evaluate the attack at $\epsilon$ of 0.031, 0.05, and 0.062, respectively, and find that the less the maximum perturbation, the lower the attack efficiency. Especially for HSJA, on ViT-B, when $\epsilon = 0.062$, the AQ and ASR are 264.23 and 91.61. When $\epsilon = 0.031$, the AQ increases slightly to 288.08, but the ASR drops significantly to 38.03. SignFlip, RayS, and Ours have high ASR at different eps, but our attack has the best attack efficiency, which shows its applicability in adversarial robustness evaluation.

\subsection{Ablation Study}
In this section, we choose ResNet50~\cite{resnet} and ViT~\cite{vit} as the threat models and randomly select 128 faces (4$\times$32=128) on FaceForensics++~\cite{ffpp} for the ablation study. 

\begin{table}[t]
\caption{Ablation study on key modules. Frequency noises and cross-task perturbation initialization significantly improve the attack efficiency, illustrating its effectiveness.}
\label{tab:cpiandnoise}
\centering
\setlength{\tabcolsep}{1.5mm}
\scalebox{0.82}{
\begin{tabular}{@{}cccccccc@{}}
\toprule
\multirow{2}{*}{Fre\_Noise} & \multirow{2}{*}{CPI} & \multicolumn{3}{c}{ResNet50} & \multicolumn{3}{c}{ViT} \\ \cmidrule(l){3-8} 
 &  & AQ$\downarrow$ & MQ$\downarrow$ & ASR(\%)$\uparrow$ & AQ$\downarrow$ & MQ$\downarrow$ & ASR(\%)$\uparrow$ \\ \midrule
 &  & 57.79 & 48.0 & 100.00 & 167.99 & 72.0 & 99.21 \\
\checkmark &  & 53.95 & 44.0 & 100.00 & 164.23 & 66.0 & 100.00 \\
 & \checkmark & 27.25 & 16.0 & 100.00 & 160.04 & 62.0 & 100.00 \\
\checkmark & \checkmark & \textbf{23.75} & \textbf{14.0} & 100.00 & \textbf{158.42} & \textbf{56.0} & 100.00 \\ \bottomrule
\end{tabular}
}
\end{table}

\textbf{Key Modules.} We first investigate the impact of the proposed key modules on the attack efficiency. Fre\_Noise represents whether to add noises in the frequency domain (refer to Eq.~\ref{fre_noise}), and CPI represents cross-task perturbation initialization. As shown in Table~\ref{tab:cpiandnoise}, adding frequency noises or cross-task perturbation initialization individually brings gains (i.e., 6.65\%/52.84\% on ResNet50, 2.23\%/4.73\% on ViT), revealing the effectiveness of the corresponding key modules. Combing all modules together leads to the best attack performance, yielding the query efficiency of 58.90\% and 5.69\% on ResNet50 and ViT, respectively.

\begin{figure}[t]
    \centering
    \includegraphics[width=0.5\textwidth]{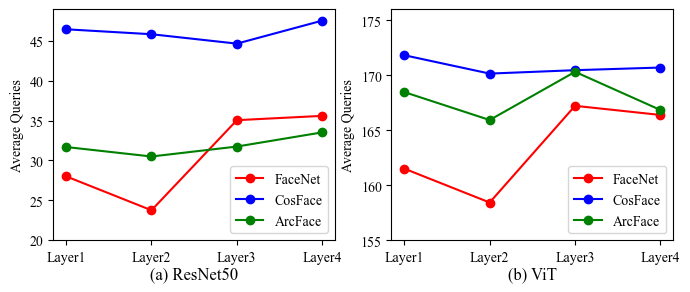}
    \caption{Ablation study on intermediate feature layers. We find that middle-layer features tend to have better attack performance. In addition, FaceNet has a better initialization effect than CPI's face recognition model.}
    \label{fig:layers}
\end{figure}

\textbf{Intermediate Feature Layers.}
We choose different face recognition models (FaceNet~\cite{FaceNet}\footnote{\url{https://github.com/ShawnXYang/Face-Robustness-Benchmark/blob/master/RobFR/networks/FaceNet.py}}, CosFace\cite{CosFace}\footnote{\url{https://github.com/ShawnXYang/Face-Robustness-Benchmark/blob/master/RobFR/networks/CosFace.py}}, and ArcFace\cite{ArcFace}\footnote{\url{https://github.com/ShawnXYang/Face-Robustness-Benchmark/blob/master/RobFR/networks/ArcFace.py}}) and consider choosing different intermediate feature layers to craft cross-task adversarial perturbations.
Table~\ref{tab:layerselection} illustrates the selection of intermediate feature layers, and Figure~\ref{fig:layers} illustrates the average number of queries for different intermediate feature layers. Attacking the features of the middle layer achieves better results than attacking the top or bottom layers, so we choose to use the bold layers in Table~\ref{tab:layerselection} to generate cross-task adversarial perturbations. Besides, FaceNet has better attack efficiency than other models, so we choose FaceNet as the face recognition model initialized by cross-task perturbation.

\begin{table}[t]
\caption{The selection of intermediate feature layers of face recognition models. The \textbf{bolded intermediate feature layers} are used to implement cross-task perturbation initialization.}
\label{tab:layerselection}
\centering
\scalebox{1}{
\begin{tabular}{@{}ccccc@{}}
\toprule
Model & Layer1 & Layer2 & Layer3 & Layer4 \\ \midrule
FaceNet & conv2d\_2b & \textbf{conv2d\_4b} & mixed\_7a & block8 \\
CosFace & layer1 & layer2 & \textbf{layer3} & layer4  \\
ArcFace & layer1 & \textbf{layer2} & layer3 & layer4 \\ \bottomrule
\end{tabular}
}
\end{table}

\begin{table}[t]
\caption{Ablation study on the frequency noise magnitude $\gamma$.}
\label{tab:gamma}
\centering
\scalebox{0.88}{
\begin{tabular}{@{}c|ccc|ccc@{}}
\toprule
\multirow{2}{*}{$\gamma$} & \multicolumn{3}{c|}{ResNet50} & \multicolumn{3}{c}{ViT} \\ \cmidrule(l){2-7} 
 & AQ$\downarrow$ & MQ$\downarrow$ & ASR(\%)$\uparrow$ & AQ$\downarrow$ & MQ$\downarrow$ & ASR(\%)$\uparrow$ \\ \midrule
2.5 & 23.00 & 14.00 & 100.00 & 171.29 & 42.00 & 100.00 \\
2.0 & 23.89 & 14.00 & 100.00 & 163.09 & 42.00 & 100.00 \\
1.75 & 23.75 & 14.00 & 100.00 & 158.42 & 42.00 & 100.00 \\
1.5 & 24.34 & 14.00 & 100.00 & 158.42 & 46.00 & 100.00 \\
1.0 & 26.15 & 14.00 & 100.00 & 152.63 & 54.00 & 100.00 \\ \bottomrule
\end{tabular}
}
\end{table}

\textbf{Frequency noise magnitude $\gamma$.} The magnitude of the frequency perturbation is critical to the efficiency of the attack. Table~\ref{tab:gamma} illustrates the effect of different frequency noise magnitudes $\gamma$ on the attack efficiency.
The hyperparameter $\gamma$ is set to 1.75 based on a comprehensive analysis of ablation study results. This value achieves an optimal balance among multiple key metrics across ResNet50 and ViT. Specifically, compared to larger $\gamma$ values (e.g., 2.0 and 2.5), a $\gamma$ of 1.75 reduces AQ for ViT from 171.29 to 158.42, indicating improved efficiency, while maintaining comparable AQ for ResNet50. Meanwhile, decreasing $\gamma$ below 1.75 (e.g., 1.5 and 1.0) increases MQ for ViT. Thus, $\gamma = 1.75$ optimizes both efficiency and stability, and demonstrates generalizability across different model architectures, making it the ideal choice for practical deployment.

\begin{figure}[!t]
    \begin{center}
    \includegraphics[width=0.47\textwidth]{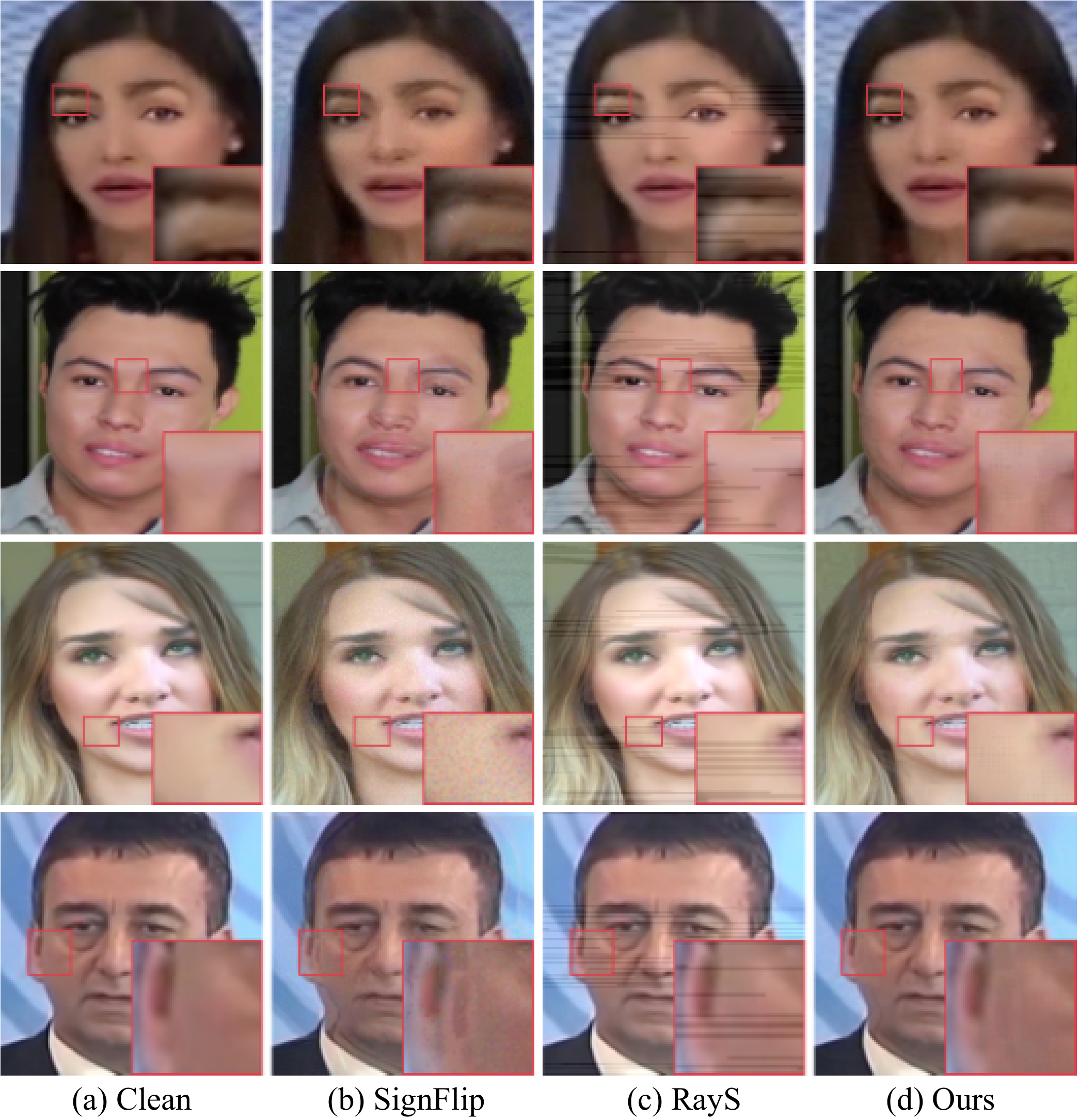}
    \end{center}
    \caption{Visual qualitative evaluation on SignFlip, RayS, and ours on FaceForensics++. From top to bottom, the forged faces are from Deepfake, Face2Face, FaceSwap, and NeuralTextures. }
    \label{fig:vis2}
\end{figure}

\subsection{Image Quality Assessment}
In this section, we analyze the image quality assessment of adversarial examples generated on FaceForensics++~\cite{ffpp}. We choose Xception as the threat model, and state-of-the-art Signflip~\cite{SignFlip} and RayS~\cite{RayS} for comparison. To assess the image quality, MSE, PSNR, and SSIM~\cite{ssim} are used as evaluation metrics to quantify the difference between adversarial and real faces. As shown in Table~\ref{tab:quality}, our method outperforms other attacks by a wide margin in terms of MSE, PSNR, and SSIM. Further, we visualize the adversarial examples generated by SignFlip, RayS, and our method in Figure~\ref{fig:vis2}. The adversarial examples generated by our method are more imperceptible, unlike SignFlip with artifacts or RayS with obvious black lines. Finally, to verify that better image quality can attack both face recognition and face forgery detection, we choose the recognition success rate (RSR, \%) as the evaluation criterion. We compare the face of adversarial examples in Table~\ref{tab:spatial} with the real face of its same identity, and the results are shown in Table~\ref{tab:quality}.
In addition, frequency noise significantly improves image quality.
It shows that our method has stronger attack performance while maintaining high image quality since our method is able to break both face recognition and face forgery detection.

\begin{table}[t]
\caption{Quantitative image quality assessment of adversarial examples generated by SignFlip~\cite{SignFlip}, RayS~\cite{RayS}, and our method on FaceForensics++~\cite{ffpp}.}
\label{tab:quality}
\centering
\scalebox{0.93}{
\begin{tabular}{@{}ccccc@{}}
\toprule
Attack & MSE$\downarrow$ & PSNR$\uparrow$ & SSIM$\uparrow$ & RSR(\%)$\uparrow$ \\ \midrule
Clean & - & - & - & \textbf{73.75} \\
SignFlip & 127.4 & 27.40 & 0.8525 & 72.67 \\
RayS & 81.4 & 29.54 & 0.7630 & 71.96 \\
Ours \textit{w.o.} Fre\_Noise & 74.4 & 30.11 & 0.8195 & 73.21 \\
Ours & \textbf{42.7} & \textbf{32.63} & \textbf{0.8766} & 73.39 \\ \bottomrule
\end{tabular}
}
\end{table}

\subsection{Attacks on industrial APIs}
\label{sec:api}
In this section, we investigate the applicability of decision-based attacks on real-world systems, such as commercial face forgery detection APIs.
We regard the face forgery detection API\footnote{\url{https://cloud.tencent.com/product/atdf?fromSource=gwzcw}} on Tencent AI Open Platform as the threat model. We choose the detection threshold to be 0.5, that is, if the output is greater than the threshold, the input is judged as fake faces. Simultaneously, we choose the face recognition API\footnote{\url{https://ai.qq.com/product/face.shtml\#\#compare}} on Tencent AI Open Platform to calculate the recognition success rate (RSR, \%). For face comparison, we choose a similarity threshold of 60. We choose 100 random frames of FaceForensics++~\cite{ffpp} as test samples, including four forging methods, Face2Face~\cite{Face2Face}, FaceSwap~\cite{Faceswap}, DeepFakes~\cite{Deepfake} and NeuralTextures~\cite{neuraltextures}.

\begin{figure}
    \centering
        \includegraphics[width=0.48\textwidth]{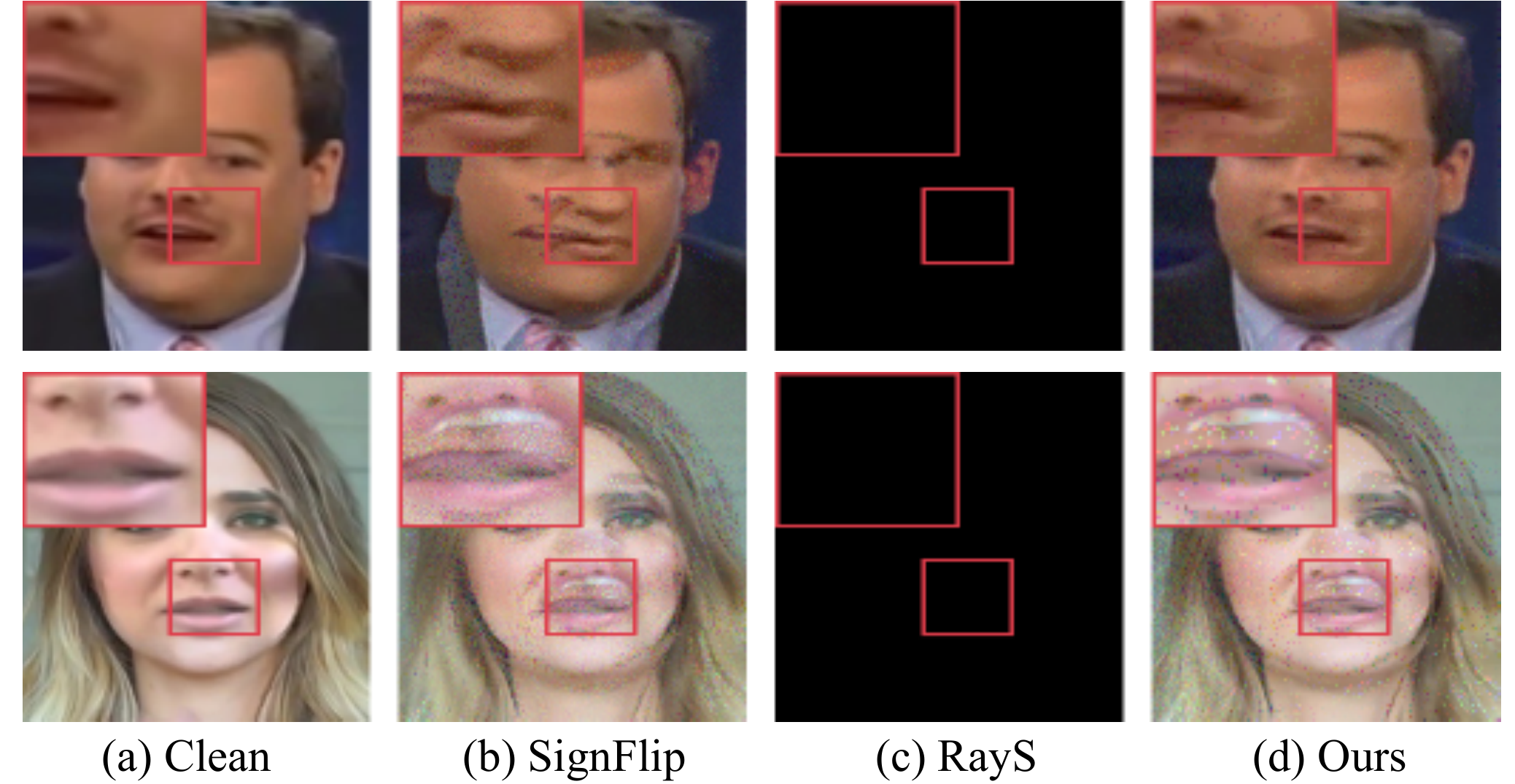}
    \caption{Visualization of generated adversarial examples for attacking commercial APIs ($\epsilon=0.125$). RayS is denied access because the perturbation initialization fails to detect faces. Compared with SignFlip, our method has a higher recognition success rate and can better attack both face recognition and face forgery detection.}
    \label{fig:api}
\end{figure}

Here, we set the maximum number of queries to $1,000$ and the maximum perturbation $\epsilon$ to 0.062 and 0.125. RSR is only calculated on successful samples identified as real faces. Table~\ref{tab:api} shows the experimental results of attacking industrial APIs, and Figure~\ref{fig:api} shows the visualization of attacking commercial APIs. Note that RayS~\cite{RayS} is ineffective in attacking industrial APIs. This is because RayS' perturbation initialization causes the image to become black, making the API unable to detect faces, and eventually leading to API denial of service. Compared with SignFlip~\cite{SignFlip}, our method has a higher attack success rate (26\% $\uparrow$ and 1\% $\uparrow$) and lower query efficiency under different $\epsilon$. 
In terms of recognition success rate (RSR), our method outperforms the baseline with an improvement of 17.00\% and 3.00\%, respectively, which can be attributed to the better image quality produced by our method.
This suggests that our method is more effective in simultaneously attacking face recognition and face forgery detection systems. Therefore, our method can effectively evaluate the security of existing AI systems in real-world scenarios.

\begin{table}[t]
\caption{Experimental results for attacking industrial APIs.}
\label{tab:api}
\centering
\setlength{\tabcolsep}{1mm}
\scalebox{0.73}{
\begin{tabular}{@{}c|cccc|cccc@{}}
\toprule
\multirow{2}{*}{Attack} & \multicolumn{4}{c|}{$\epsilon=0.062$} & \multicolumn{4}{c}{$\epsilon=0.125$} \\ \cmidrule(l){2-9} 
 & AQ$\downarrow$ & MQ$\downarrow$ & ASR(\%)$\uparrow$ & RSR(\%)$\uparrow$ & AQ$\downarrow$ & MQ$\downarrow$ & ASR(\%)$\uparrow$ & RSR(\%)$\uparrow$ \\ \midrule
SignFlip & 92.80 & \textbf{12.0} & 40.00 & 21.00 & 39.51 & \textbf{12.0} & 74.00 & 24.00 \\
RayS & - & - & 0.00 & 0.00 & - & - & 0.00 & 0.00 \\
Ours & \textbf{80.18} & \textbf{12.0} & \textbf{66.00} & \textbf{38.00} & \textbf{38.18} & \textbf{12.0} & \textbf{75.00} & \textbf{27.00} \\ \bottomrule
\end{tabular}
}
\end{table}

\subsection{Adversarial Transferability}
Our approach is query-based, achieving an approximate 100\% attack success rate with exceptionally high query efficiency. To further enhance our evaluation, we assess the cross-dataset adversarial transferability of various decision-based attacks. Specifically, we generate adversarial examples on the CelebDF dataset~\cite{CelebDF} and subsequently test the same model on the FFDF dataset~\cite{ffpp}. Table~\ref{tab:cross} illustrates the attack success rate (\%) in this cross-dataset evaluation. Our method demonstrates superior cross-dataset adversarial transferability due to its ability to mitigate differences between real and fake faces in the frequency domain.

\begin{table}[t]
\caption{Cross-dataset evaluation on adversarial transferability.}
\label{tab:cross}
\centering
\scalebox{0.9}{
\begin{tabular}{@{}ccccc@{}}
\toprule
Attack & Clean & SignFlip & RayS & Ours \\ \midrule
ResNet50 & 8.8 & 50.8 & - & \textbf{59.6} \\
ViT & 17.8 & 80.4 & 54.2 & \textbf{81.0} \\ \bottomrule
\end{tabular}}
\end{table}

\subsection{JPEG Defense}
JPEG~\cite{jpeg} is a lossy image compression method designed to preferentially preserve details important to the human visual system. The human eye is more sensitive to low-frequency components and less sensitive to high-frequency details. JPEG compression takes advantage of this visual characteristic and can significantly reduce the file size while maintaining the visual quality of the image by removing or reducing the amount of images of high-frequency components. Therefore, it has become a mainstream adversarial defense method to eliminate noises. To further evaluate the robustness of our attack, we evaluate JPEG defenses~\cite{jpeg} under different compression rates. As shown in Table~\ref{tab:jpeg}, a stronger compression rate reduces the detection accuracy of the image.
Since JPEG~\cite{jpeg} removes high-frequency features, our FDA adds noise at both high and low frequencies, thus keeping state-of-the-art attack performance because low-frequency noise is still effective. In addition, the attack performance of SignFlip and RayS is degraded because they do not consider the frequency noise, which shows their shortcomings in robustness evaluation.

\begin{table}[t]
\caption{Attack performance with the JPEG defense on FFDF.}
\label{tab:jpeg}
\begin{center}
\scalebox{0.9}{
\begin{tabular}{@{}ccccc@{}}
\toprule
\multirow{2}{*}{Attack} & \multirow{2}{*}{w.o. defense} & \multicolumn{3}{c}{JPEG Compression Rate} \\ \cmidrule(l){3-5} 
 &  & 0.5 & 0.7 & 0.9 \\ \midrule
Clean & 0.00 & 24.82 & 6.61 & 0.17 \\
SignFlip & \textbf{100.00} & 92.14 & 83.21 & 86.07 \\
RayS & \textbf{100.00} & \textbf{95.16} & 93.38 & 81.75 \\
Ours & \textbf{100.00} & \textbf{95.16} & \textbf{95.54} & \textbf{95.00} \\ \bottomrule
\end{tabular}
}
\end{center}
\end{table}

\section{Discussion}
\label{sec:discussion}
In this section, we further analyze the activation relationship of Cross-task Perturbation Initialization in face recognition and face forgery detection features in Section~\ref{sec:cpi}. Face recognition distinguishes the identities of different faces. Recent work~\cite{zhuang2022towards,yu2023diff} focuses on using face identities to enhance face forgery detection. The middle layer features of face recognition are positively correlated with face forgery detection, so the adversarial examples of face recognition can be used to attack face forgery detection, and we propose Cross-task Perturbation Initialization. Our work is the first to simultaneously attack two face-related tasks, face forgery detection and face recognition, in academic and industrial scenarios. Cross-task Perturbation Initialization on face recognition is also beneficial to attack face recognition and face forgery detection. Therefore, the introduction of face recognition may be the future research direction of face forgery detection.

\noindent\textbf{Broader impacts}.
The method proposed in this paper may threaten face-related applications to some extent, such as face recognition and face forgery detection. Further, through attacks, the method can mine the security vulnerabilities of existing applications, which is beneficial to the development of more secure applications in the future.

\section{Conclusions}
\label{sec:conclusions}
In this paper, we first explore decision-based attacks on face forgery detection. The direct application of existing decision-based attacks, however, suffers from initialization failures and low image quality. To alleviate initialization sensitivity on attacks, we propose cross-task perturbation that utilizes the high correlation of face features across tasks. Based on the frequency cues utilized in face forgery detection, we propose a frequency decision-based attack. In particular, we perturb the frequency domain and then constrain the visual quality in the spatial domain. Our method achieves high query efficiency and guaranteed image quality for FaceForensics++, CelebDF, and industrial APIs, demonstrating state-of-the-art attack performance and exposing the adversarial vulnerability of existing detectors. We hope that this method can better evaluate the adversarial robustness of face forgery detection.

\section*{CRediT authorship contribution statement}
Zhaoyu Chen: Conceptualization, Investigation, Experimentation, Visualization, Writing-Original Draft, Review \& Editing.
Bo Li: Supervision, Writing-Original Draft, Review \& Editing.
Kaixun Jiang: Experimentation, Review \& Editing. 
Shuang Wu: Investigation, Review \& Editing.
Shouhong Ding: Review \& Editing
Wenqiang Zhang: Supervision, Review \& Editing, Funding acquisition
\section*{Declaration of competing interest}
The authors declare that they have no known competing financial
interests or personal relationships that could have appeared to influence
the work reported in this paper. 
\section*{Data availability}
Data will be made available on request. 
\section*{Acknowledgments}
This work was supported by National Natural Science Foundation of China (No.62072112), Scientific and Technological innovation action plan of Shanghai Science and Technology Committee (No.22511102202), Fudan Double First-class Construction Fund (No. XM03211178). This work was completed by Zhaoyu Chen during his internship at Youtu Lab, Tencent.




\bibliographystyle{model5-names}

\bibliography{cas-refs}

\end{document}